\documentclass[nohyperref]{article}
\usepackage{microtype}
\usepackage{graphicx}
\usepackage{subfigure}
\usepackage{pifont}
\usepackage{amssymb}
\usepackage{booktabs}
\newcommand{\ra}[1]{\renewcommand{\arraystretch}{#1}}
\usepackage{hyperref}

\usepackage[accepted]{icml2022}
\usepackage{amsmath}
\usepackage{amssymb}
\usepackage{mathtools}
\usepackage{amsthm}
\usepackage[capitalize,noabbrev]{cleveref}

\theoremstyle{plain}

\theoremstyle{definition}

\theoremstyle{remark}

\usepackage[textsize=tiny]{todonotes}

\icmltitlerunning{~ \hfill  Mixed Precision Quantization with Sensitivity Guided Search \hfill \thepage }

\begin{document}

\twocolumn[
\icmltitle{Mixed Precision Post Training Quantization of Neural Networks with Sensitivity Guided Search}

\begin{icmlauthorlist}
\icmlauthor{Clemens JS Schaefer}{sch}
\icmlauthor{Elfie Guo}{comp}
\icmlauthor{Caitlin Stanton}{comp}
\icmlauthor{Xiaofan Zhang}{comp}
\icmlauthor{Tom Jablin}{comp}
\icmlauthor{Navid Lambert-Shirzad}{comp}
\icmlauthor{Jian Li}{comp}
\icmlauthor{Chiachen Chou}{comp}
\icmlauthor{Siddharth Joshi}{sch}
\icmlauthor{Yu Emma Wang}{comp}
\end{icmlauthorlist}

\icmlaffiliation{comp}{Google LLC, Mountain View, CA, USA}
\icmlaffiliation{sch}{University of Notre Dame, Department of Computer Science and Engineering, Notre Dame, IN, USA}

\icmlcorrespondingauthor{Clemens JS Schaefer}{cschaef6@nd.edu}
\icmlcorrespondingauthor{Emma Wang}{yuemmawang@google.com}

\icmlkeywords{Quantization, Mixed Precision Quantization, Post Training Quantization, PTQ, Hessian, Quantization Error, Noise Injection, Bit Allocation, Bit Configuration, Quantization Configuration Search, ResNet50, BERT, ImageNet, SQuAD}

\vskip 0.3in
]

\printAffiliationsAndNotice{}

\begin{abstract}
Serving large-scale machine learning (ML) models efficiently and with low latency has become challenging owing to increasing model size and complexity. Quantizing models can simultaneously reduce memory and compute requirements, facilitating their widespread access. However, for large models not all layers are equally amenable to the same numerical precision and aggressive quantization can lead to unacceptable loss in model accuracy. One approach to prevent this accuracy degradation is mixed-precision quantization, which allows different tensors to be quantized to varying levels of numerical precision, leveraging the capabilities of modern hardware. Such mixed-precision quantiztaion can more effectively allocate numerical precision to different tensors `as needed' to preserve model accuracy while reducing footprint and compute latency. In this paper, we propose a method to efficiently determine quantization configurations of different tensors in ML models using post-training mixed precision quantization. We analyze three sensitivity metrics and evaluate them for guiding configuration search of two algorithms. We evaluate our method for computer vision and natural language processing and demonstrate latency reductions of up to $27.59\%$ and $34.31\%$ compared to the baseline 16-bit floating point model while guaranteeing no more than $1\%$ accuracy degradation.
\end{abstract}

\section{Introduction}

\begin{figure}[ht!]
\begin{center}
\centerline{\includegraphics[width=.85\columnwidth, trim={0.4cm 0.2cm 0.3cm 0.4cm},clip]{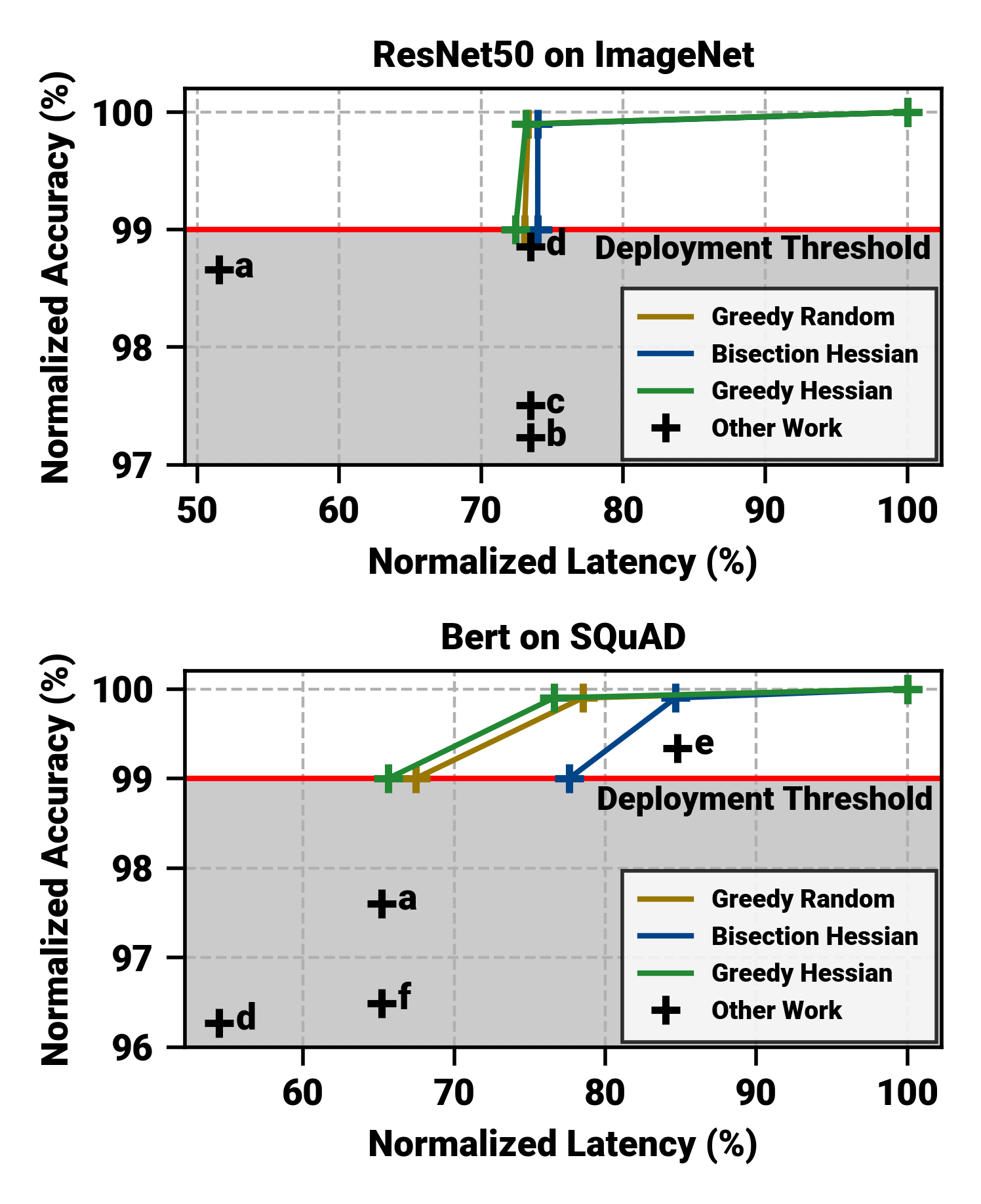}}
\vskip -0.2in
\caption{Overview of our results in comparison to relevant prior work. Axis are normalized to the performance of a 16 bit floating point model. The shaded region indicates when model accuracy degrades below thresholds acceptable for production deployment ($\gg1\%$~\cite{jouppi2021ten}). The curves show the performance of our methods given an accuracy target of 99\% and 99.9\% as well as the the floating point model performance (100\% on both axis). Demonstrating the superiority of our Hessian-based greedy search with the random based greedy search in close proximity. Letters indicate results from related prior work: \textbf{a} \citet{hubara2021accurate}, \textbf{b} \citet{nahshan2021loss}, \textbf{c} \citet{nagel2020up}, \textbf{d} \citet{wu2020integer}, \textbf{e} \citet{shen2020q}, \textbf{f} \citet{jeon2022mr}, }
\label{fig:over}
\end{center}
\vskip -0.3in
\end{figure}

Neural networks (NNs) are the driving force behind a long series of breakthroughs in artificial intelligence, delivering state-of-the-art performance across a wide range of tasks, most notably in computer vision~\cite{zhai2022scaling}, natural language processing~\cite{brown2020language} and generative models for text~\cite{thoppilan2022lamda} or images~\cite{ramesh2021zero}. However, the remarkable capabilities of these state-of-the-art (SOTA) NNs incur substantial compute and memory costs, making them challenging to deploy at scale. These costs are additionally compounded by machine learning (ML) proliferating across a wider range of applications~\cite{jumper2021highly}. As a result, there is a need for ML models that can be practically deployed at low-cost while meeting stringent quality of service (QoS) metrics (e.g., latency and accuracy).

\begin{figure*}[ht]
\begin{center}
\centerline{\includegraphics[width=.95\textwidth, trim={0 0 1.9cm 0},clip]{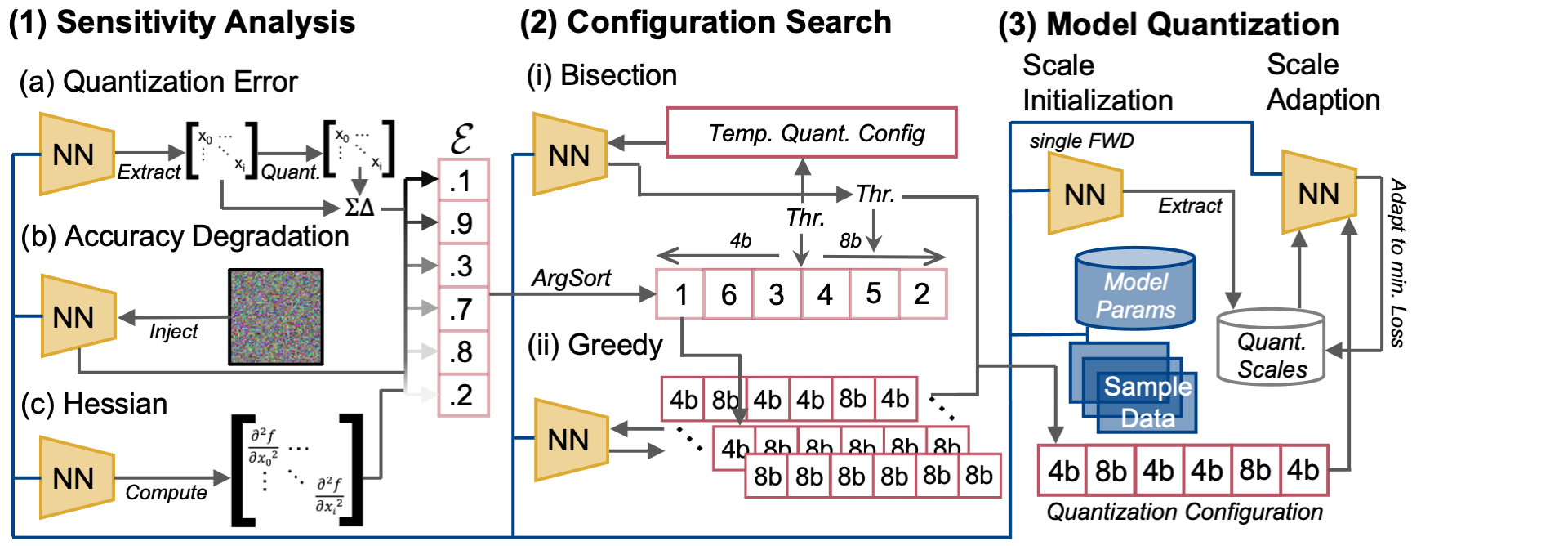}}
\vskip -0.1in
\caption{Summary of our approach. On the left we show an illustration of the sensitivity metrics: (a) quantization error which extracts floating-point matrices from the network and quantizes them, (b) noise accuracy degradation which evaluates noise injected NNs and (c) second derivative information which computes the Hessian matrix of individual layers. In the middle of the Figure we sketch conceptually our two bit-width search algorithms i) bisection search and ii) greedy search. And on the right we present our PTQ approach, which uses calibrates the quantization scales with the maximum value first and then adapts them using overall network loss minimization. }
\label{fig:meth}
\end{center}
\vskip -0.4in
\end{figure*}

To fulfil this need and support the current pace of innovation and scaling of machine learning models, researchers have proposed various methods to reduce the computational cost of NNs, such as hardware-efficient NN designs~\cite{tan2019efficientnet}, pruning~\cite{mishra2021accelerating} and quantization~\cite{gholami2021survey}. Among these, quantization can be model agnostic and be used in conjunction with other methods. By reducing the precision (bit-widths) of the weights and activations of a NN, quantization significantly reduces the memory footprint and enables the use of cheaper compute primitives, thereby increasing the NN inference speed (latency reduction) and reducing the compute-energy costs.

Other work in literature mitigates quantization induced accuracy degradation by additionally training NN models to compensate for reduced precision, typically referred to in literature as quantization aware training (QAT)~\cite{dong2020hawq, wang2019haq}. However, QAT requires substantial amounts of training data and necessitates profound changes to the model parameters, which might have unintended consequences when a model is deployed. Alternatively, post-training quantization (PTQ) approaches determine adequate quantization scales on small calibration data sets while minimizing the change to model parameters. This trades-off quantization complexity and model-revalidation against the model accuracy~\cite{wu2020integer}.

Additionally, modern hardware such as Google TPUs~\cite{jouppi2021ten} or NVIDIA A100~\cite{nvidia2020a100} support quantized operations at various bit-widths, e.g. int4, int8, fp8, fp16, fp32 or fp64, to facilitate efficient NN inference. However, to exploit these hardware capabilities in practice, the different NN layers and operations must be performed at (or configured to) an optimal bit-width which balances model accuracy with efficiency. Since the search space of all possible bit-width (quantization) configurations is exponential with the number of layers, this represents a further challenge for optimal quantization. QAT tackles that challenge by: i) making the bit width a learnable parameter and training the bit width alongside other model parameters (given model size constraints)~\cite{schaefer2022edge, uhlich2019mixed}, ii) using unexplainable black-box reinforcement learning solutions to determine bit-widths~\cite{wang2019haq}, or iii) using an auxiliary metric to reduce the search space~\cite{dong2020hawq}. Owing to the increased complexity encountered in PTQ, most PTQ approaches have only focused model-wide single-bit-width quantization and avoided finer-grained bit-width allocation.

In this work we develop a method for efficient PTQ that facilitates deploying floating-point machine learning models with minimal manual intervention (see Figure~\ref{fig:meth}) on recent hardware with multi-precision support. We do so through the following contributions:

\begin{itemize}
\item \textbf{Quantization Sensitivity Metrics}: We conduct an in-depth analysis of three metrics to guide the quantization search quantization error, noise performance degradation, and second order information. 
\item \textbf{Guided Precision Configuration Search}: We propose two sensitivity-guided search algorithms (bisection and greedy) to identify optimal network quantization configurations which maintain a minimum accuracy level for PTQ.
\item \textbf{Comprehensive Evaluation}: We evaluate our technique and experimentally show improvements over SOTA PTQ for model size reduction and latency on both a convolutional vision model and a transformer-based language model. Experimental results show latency reductions of up to $27.59\%$ (ResNet50) and $34.31\%$ (BERT), while outperforming prior work (Figure~\ref{fig:over}).
\end{itemize}

\section{Related Work}

There have been interesting and significant efforts to increase PTQ accuracies, such as ~\citet{hubara2021accurate} (\textbf{a} in Fig.~\ref{fig:over}) proposed a quantization method which updates model parameters to minimizes the error between the quantized layer output and full precision output and also fine-tune batchnorm parameters. Furthermore, they introduce a integer programming formulation to optimally allocate the bit-widths for each layer. This method changes the model weights as well as batchnorm parameters and can not guarantee QoS, thereby potentially complicating deployment and putting it on the border between PTQ and QAT. Coming from an analytical angle \citet{nahshan2021loss} (\textbf{b} in Fig.~\ref{fig:over}) investigate the effect of quantization on the structure of the loss landscape and found flat separable structures for mild quantization and highly non-separable with steep curvature for low bit-width quantization. Inspired by their analysis they devise a three step method to improve PTQ for lower bit-widths: i) determine the quantization step that minimizes a norm of the quantization error of the individual layers, ii) use quadratic interpolation to approximate an optimum quantization scale, and iii) jointly optimize the parameters of all layers acquired on the previous step by applying a gradient-free optimization method. In a similar way \citet{nagel2020up} (\textbf{c} in Fig.~\ref{fig:over}) theoretically analyze the rounding decision of the quantization process and view the rounding decision (up or down) as a binary optimization problem. As a solution they propose a layer-wise local loss, which can be optimized with a soft relaxation, determining the rounding decision and improving PTQ accuracies. In a larger empirical study~\citet{wu2020integer} (\textbf{d} in Fig.~\ref{fig:over}) evaluated various calibration techniques for PTQ and found that overall entropy based and percentile calibrations work well however the exact setting of the percentile calibration is model and data dependent.

Mixed precision quantization approaches need to determine how many bits to allocate to each layer. \citet{wang2019haq} use reinforcement learning to automatically determine the quantization policy using feedback from a hardware accelerator. They demonstrate reductions of latency by $1.4-1.95\times$ and the energy consumption by $1.9\times$ compared to eight bit integer models with comparable accuracy. Other works~\cite{schaefer2022edge} extend QAT to also train the bit-width of each layer along the model parameters and demonstrate the effectiveness of their method by extending the SOTA pareto frontier with models sized below 4.3MB for the ImageNet task. Similarly~\citet{jeon2022mr} also consider the quantization parameters, e.g. scaling factors and bit-codes, as learnable and jointly optimize them with the weights of transformer models (\textbf{f} in Fig.~\ref{fig:over}).  Another approach is to determine layer importance or sensitivities and allocate bit widths accordingly. This so far has not been done for PTQ, however some works have applied this idea to QAT methods. For example~\citet{dong2020hawq} propose to use the mean of the Hessian trace to determine layer sensitivity and subsequently employ a Pareto-frontier based technique to determine good quantization configurations. They reduce the size of a ResNet50 to 7.99MB which achieves 75.76\% accuracy. Equivalently~\citet{shen2020q} use a group-wise quantization scheme and Hessian based mix-precision method on BERT (\textbf{e} in Fig.~\ref{fig:over}).~\citet{yang2021mixed} present a dynamical Hessian matrix based mixed-precision quantization scheme specifically for object detection, which iteratively selects a layer with the lowest sensitivity based on the Hessian matrix and downgrades its precision. Additionally they use a loss function that takes into account the joint effects of quantization on classification and regression loss. On the COCO task with a RetinaNet the scheme achieves 37.1\% mean average precision with an average bit-width 3.99.

\section{Methods}

\subsection{Quantization}

Fixed point quantization often termed integer quantization  reduces the precision of numerical values in a model for a resulting decrease in storage and compute requirements. This is typically achieved by applying clipping and rounding operations to the original floating-point values, often formulated as:
\begin{equation}
Q(\mathbf{x}) = \text{round}(\text{clip}(\alpha \cdot \mathbf{x} ) \cdot 2^{b-1}) \cdot 2^{-(b-1)} \cdot \gamma.
\end{equation}
Here, $Q$ is the quantization function, $\mathbf{x}$ is the floating point value, the clipping function clips the value between -1 and 1, $b$ is the bit width, and $\alpha$ and $\gamma$ are the quantization scales. While the majority of literature employs a single quantization scale ($\gamma = \alpha^{-1}$). Recent work has demonstrates superior results from employing two scaling parameters, to better align the pre- and post-quantized values~\cite{park2020profit}. As part of PTQ determining the optimal values of these scale parameters need to be determined to minimize model accuracy degradation. 

Quantization scales are estimated using a two-step process (Figure~\ref{fig:meth} \textbf{(3)}). The first step (calibration) uses a calibration data set and calculates the maximum value in each  tensor in the model (e.g., weights and activations). We then set $\alpha = 1/\max(x)$ and $\gamma = \max(x)$), where $x$ is the tensor of interest. In the second step (adjustment), we adapt the scales using backpropagation to minimize the model loss on the calibration dataset. Since this only adapts scales, we do not update the parameters of the original model, facilitating easy deployment.

\subsection{Sensitivity Measures}

The space of possible tensor-precisions for a model is exponential with the number tensors. Consider a ResNet50 with three different configuration options just for the parameters (e.g. bit widths), this results in $3^{50}$ possible quantization configurations. Evaluating all of these configurations, even with parallelization, represents a substantial computational workload. Consequently searching through this space efficiently is critical to deploying performant quantized models. The use of an informative sensitivity metric can reduce this vast space, making it practical to search for valid and performant configurations. Such a metric should encapsulate how the model accuracy might be impacted by quantizing different layers to inform the quantization process.

\subsubsection{Quantization Error ($\mathcal{E}_{\text{QE}}$)}

Quantization error (QE) forms a natural baseline sensitivity metric against which more complex metrics might be evaluated. We use the root-mean-square error, between a quantized and unquantized tensor, normalized by the maximum absolute value of the tensor as the QE for this tensor ($\mathcal{E}_{QE}$). This can be written as:
\begin{equation}
\mathcal{E}_{\text{QE}} =\sqrt{ \mathbf{E}[(Q(\mathbf{x}) - \mathbf{x})^2]} \cdot \frac{1}{\max |\mathbf{x}|}.
\end{equation}
While QE can directly capture the impact of reduced precision operation for a single tensor, it may not fully capture the impact of quantization on the entire model. Consider the case where high QE at certain layers does not degrade model performance, while a comparatively small QE at other layers may have a larger impact. We refer the reader to Figure~\ref{fig:sens} in the appendix for a concrete example.

\subsubsection{Accuracy Degradation from Noise ($\mathcal{E}_{\text{N}}$)}
% scale half the calibration data set
As an alternative to using QE, a tensor's resilience to perturbations might serve as a viable estimation of the tensor's sensitivity within the model. We inject Gaussian noise into a single tensor in the model and assess the model's performance on the calibration data set. We employ a Gaussian noise model building on related work~\citet{park2022nipq} which observed superior performance for Gaussian noise as compared to the more conventionally used uniform distribution. We define the sensitivity metric ($\mathcal{E}_{\text{N}}$) as the difference in performance between the noise free model and noise injected model, expressed as:
\begin{align}
    \mathcal{E}_{\text{N}} &= L(\mathbf{x}, \mathbb{W}^*) - L(\mathbf{x}, \mathbb{W}),\\
    \mathbb{W}^* &= \left\{\mathbb{W} \setminus{\mathbf{w_i}}, \mathbf{w_i} + \mathbf{\nu}\right\},\\
    \mathbf{\nu} &= \mathcal{N}(0, \lambda \max{|\mathbf{w_i}|}).
\end{align}
Here, the first part of the equation represents the loss $L$ of the model given the calibration data $\mathbf{x}$ and the set of all weights including one perturbed weight ($\mathbb{W}^*$) and the second part represents the loss of a model without any perturbed weights ($\mathbb{W}$). $\mathbb{W}^*$ is defined as the set of all weights ($\mathbb{W}$), with the exclusion of a single tensor ($\mathbf{w_i}$), and the inclusion of a perturbed tensor ($\mathbf{w_i}+\nu$). Since perturbations are Gaussian distributed, they can be written as a Normal distribution $\mathcal{N}$ with $0$ mean and a variance which is scale by an additional parameter $\lambda$. This same formulation can be extended to other metrics such as model accuracy and other tensors in the model such as activations.

As formulated,  $\mathcal{E}_{\text{N}}$, assumes that the un-perturbed model performance will be higher than the performance after any perturbations. However, the limited discriminatory power of this metric can impact ordering the tensors by sensitivity, a critical requirement for successful mixed-precision mode; quantization.

\subsubsection{Second Derivative Information ($\mathcal{E}_{\text{Hessian}}$)}

We investigate second-order derivative information as a sensitivity measure, which pertains to the local curvature of a function. This choice is informed by theory that model accuracy is robust to perturbations in values that occupy flat regions of the loss function (low local curvature). However, for those values that occupy regions of high local curvature (sharp), small perturbations can have an exaggerated impact on model accuracy~\cite{dong2019hawq, rissanen1978modeling, hochreiter1997flat}. One way of estimating the local curvature, uses the Hessian of the loss function, which comprises second-order partial derivatives of the loss.

Rather than directly evaluating the Hessian, which is computationally prohibitive, we approximate the Trace using Hutchinson's algorithm as seen in related work~\citet{dong2020hawq,lee2021network}. By discarding the off-diagonal elements, we trade off computational tractability against capturing cross-tensor interactions. We define a Hessian-based metric:
\begin{equation}
\mathcal{E}_{\text{Hessian}} = \mathbf{E} \Big[\mathrm{Tr} \Big[\frac{L(\mathbf{x}, \mathbb{W})}{\partial \mathbf{w}_i^2}\Big]\Big].
\end{equation}
Where $\mathrm{Tr}$ is the trace operator, $L$ the loss function of the model, $\mathbb{W}$ the set of all weights (this set could also be extended to the activations) and $\mathbf{x}$ calibration data. Larger values of $\mathcal{E}_{\text{Hessian}}$ indicate higher local curvature of the loss function and consequently greater sensitivity of the model's accuracy to lower-precision. 

% \subsubsection{Visualization of Sensitivty Metrics}

\subsection{Search for Quantization Configurations}

These sensitivity metrics are not equally informative, in guiding bit-configuration search for multi-precision PTQ. Prior work has employed such sensitivity metrics to enable a partial or exhaustive search within the configuration space to develop a pareto frontier of model-performance~\cite{dong2020hawq}. Other efforts at PTQ have avoided configuration search entirely~\cite{lee2021data}. Here, we propose to use guided search algorithms to determine per-tensor precisions for a model that can  maintain a target accuracy while minimizing latency and model footprint.

We propose two methods: a guided bisection search and a greedy approach. Both methods are progressive, starting with a floating-point model and gradually quantizing more or fewer layers based on the sensitivity metric. We iteratively reduce the bit widths of previously quantized layers to find the best configuration by using all available bit-widths. This approach is motivated by the observation that tensors insensitive to lower numerical precision can first be evaluated at higher precision, to cheaply encapsulate limited effects of cross-layer interactions. This can then be used to further shrink the search space for tensors insensitive to lower-precision quantization. 
\subsubsection{Bisection Search}
\begin{algorithm}[tb]
   \caption{Bisection search for ideal quantization configuration. Worst and average time complexity is $\mathcal{O}(b\log{N})$ with $b$ as the number of bit width choices and $N$ the number of layers.}
   \label{alg:bisect}
\begin{algorithmic}[1]
   \STATE {\bfseries Input:} data $x$, sensitivity metric $s$, accuracy  target $t$, available bit widths $bs$, model $f$.
   \STATE Initialize working configuration $w$ with $\max(bs)$.
   \STATE Initialize layer list $ll$ with all layers of $f$.
   \STATE Sort $ll$ by $s$ in ascending order.
   \FOR{$b$ {\bfseries in} $bs$}
      \STATE Initialize threhsold $thr = \text{length}(ll)/2$.
      % \STATE $noChange = false$
      \STATE Initialize upper limit $upl$ to $\text{length}(ll)$.
      \STATE Initialize lower limit $lowl$ to $0$.
      \REPEAT
        \STATE Initialize local working config $lw$ with $w$.
        \STATE $lw[ll[0:thr]] \gets b$.
        \STATE Evaluate $f(x, lw)$ and save accuracy $a$.
        \IF{$a >= t$ }
            \STATE $lowl \gets thr$.
            \STATE $thr \gets thr+(upl - thr)/2$.
        \ELSE{}
            \STATE $upl \gets thr$.
            \STATE $thr \gets thr - (thr - lowl)/2$.
        \ENDIF
      \UNTIL{$thr$ is not changing.}
      \STATE $w[ll[0:thr]] \gets b$.
      \STATE $ll \gets ll[0:thr]$. % reduce layer list for next iterations 
   \ENDFOR
   \STATE {\bfseries Return:} optimal working configuration $w$.
\end{algorithmic}
%\vskip -.2in
\end{algorithm}

The bisection search is a well known root-finding method which we apply here to determine the ideal quantization configuration. We assume that there exists a threshold sensitivity value, above which layers cannot be quantized or can only be quantized to a limited bit width. To find this threshold, the bisection search iteratively quantizes subsets of layers, where ordering is determined by a sensitivty metric. We start with a configuration which quantizes the least-sensitive half of the set of tensors. We evaluate the model with that configuration and compare the accuracy against the accurarcy target. The bisection search then proceeds to iteratively update the threshold value (and thereby the quantization configuration) by either increasing the number of layers when the accuracy target is met or decreasing the amount of quantized layers when the accuracy criteria is not fulfilled. Once the threshold sensitivity value is identified for the highest quantized precision, the method is repeated for each lower precision setting, in sequence.  The method outlined with pseudo in Algorithm~\ref{alg:bisect}. The worst and average time complexity of this search is $\mathcal{O}(b\log{N})$, where $N$ is the total number of layers and $b$ the number of available bit widths. Bisection relies on accurate ordering to reduce the search space. Consequently, inaccuracies in the ordering due to sensitivity estimations negatively impact final model performance.

\subsubsection{Greedy Approach}

\begin{algorithm}[tb]
  \caption{Greedy approach for ideal quantization configuration. Average time complexity is $\mathcal{O}((2-2^{-(b-1)})N)$ and worst case $\mathcal{O}(bN)$ where $b$ is the number of bit width choices and $N$ the number of layers. }
  \label{alg:greedy}
\begin{algorithmic}[1]
  \STATE {\bfseries Input:} data $x$, sensitivity metric $s$, accuracy target $t$, available bit widths $bs$, model $f$.
  \STATE Initialize working configuration $w$ with $\max(bs)$.
  \STATE Initialize layer list $ll$ with all layers of model.
  \STATE Sort $ll$ by $s$ in ascending order.
  \FOR{$b$ {\bfseries in} $bs$}
        % \STATE Initialize unquantizable layer $uql = \emptyset$  
        \STATE Initialize quantizable layer $ql \gets \emptyset$.
        \FOR{$l$ {\bfseries in}  $ll$}
            \STATE $w[l] \gets b$.
            \STATE Evaluate $f(x, w)$ and save accuracy $a$.
            \IF{$a >= t$ }
                \STATE Append $l$ to $ql$.
            \ELSE{}
                \STATE Set $w[l]$ back to last working value.
                % \STATE Append $l$ to $uql$.
            \ENDIF
        \ENDFOR
        \STATE $ll \gets ql$.
  \ENDFOR 
  \STATE {\bfseries Return:} optimal working configuration $w$.
\end{algorithmic}
\end{algorithm}
Since most metrics can only estimate sensitivity, alternative search techniques that are less sensitive to exact ordering can deliver higher final performance.  We examine a guided greedy search to progressively build up an efficient ideal quantization configuration. We first initialize a working configuration to the baseline precision for all layers (usually floating point 16). We then iterate through each layer and evaluate it at a lower precision. If the model evaluation falls below the accuracy threshold, the layer is no longer considered a candidate for further quantization. We iterate across the different bit-widths until all eligible layers have been evaluated at the lowest precision. A pseudo code implementation of this is outlined in Algorithm~\ref{alg:greedy}. The average time complexity is $\mathcal{O}((2-2^{-(b-1)})N)$ and the worst-case time complexity is $\mathcal{O}(bN)$ where $b$ is the number of bit width choices and $N$ is the number of layers. As with typical unstructured discrete optimization problems, there is no guarantee of arriving at the optimal quantization configuration. In part, due to relaxed assumptions on cross-tensor interactions.

\section{Experiments}

We evaluate our proposed methods on the ImageNet~\cite{ILSVRC15} and SQuaAD~\cite{2016arXiv160605250R} datasets, using ResNet50~\cite{he2016deep} and BERT~\cite{devlin2018bert} respectively. These are commonly accepted dataset-model combinations from the MLPerf inference suite~\cite{reddi2020mlperf}.\footnote{\url{https://mlcommons.org/}} Because calibration and determining the sensitivity requires some data, we randomly sample 512 examples from the original training set to obtain sensitivity metrics and resample another 512 examples to calibrate and adjust the quantizers. We use a learning rate of $1\times10^{-5}$ for quantization scale adaptation with the batch size for models determined by memory considerations (128 for ResNet50 and 48 for BERT). Configuration search uses the complete validation set to assess quantization efficacy.

\paragraph{Compute Latency Estimates}
Most deep learning frameworks such as TensorFlow\cite{tensorflow2015-whitepaper}, JAX~\cite{jax2018github}, or PyTorch~\cite{paszke2019pytorch} support quantization to 8-bit integers. More aggressive quantization, such as 4-bit integers, while supported in hardware (e.g., NVIDIA A100~\cite{nvidia2020a100}) have limited software support. In practice, modeling the benefits of executing quantized kernels in hardware is not trivial. This is in part due to complex interactions between memory hierarchy, bus-speeds, compute-utilization, and compiler optimizations. We capture these interactions by benchmarking the performance of key kernels such as \texttt{gemm} and \texttt{conv2d} across different numerical precisions on A100 GPUs, run for inference batch-size of one. The best performing kernels for a given tensor shape and precision were determined using the CUTLASS profiler and optimizer~\cite{Kerr_CUTLASS_2022}. We then developed estimates for model deployment latencies by suitable combination of kernel latencies for different multi-precision models.
\paragraph{Baselines}
\begin{table}[t]
\ra{1.3}
\caption{Baseline percentage-accuracy, model size in megabyte and latency in milliseconds for models quantized to the same bit-width. \textsuperscript{\ddag} Percentage below each row indicate the relative size compare to the numbers from the floating point 16 model. The accuracy results of the floating point 16 model meet MLperf requirements~\cite{reddi2020mlperf}. \textsuperscript{\textasteriskcentered} indicates integer format and \textsuperscript{\textdagger} stands for floating point.}
\label{tab:base}
\begin{center}
\begin{small}
\begin{tabular}{@{}rccc@{}}
\toprule
& \normalfont{4 Bits\textsuperscript{\textasteriskcentered}} & 8 \normalfont{Bits\textsuperscript{\textasteriskcentered}} & \normalfont{16 Bits\textsuperscript{\textdagger}}\\
\midrule
\multicolumn{1}{@{}l@{}}{\normalfont{Model = ResNet50}} & & & \\
\normalfont{Accuracy (\%)} & $0.10$ & $76.60$ & $76.93$\\
\normalfont{\textsuperscript{\ddag}Relative} & $0.13\%$ & $99.57\%$ & $100.00\%$ \\[4pt]
\normalfont{Size (MB)} & $12.75$ & $25.50$ & $51.00$ \\
\normalfont{\textsuperscript{\ddag}Relative} & $25.00\%$ & $50.00\%$ & $100.00\%$\\[4pt]
\normalfont{Latency (ms)} & $2.68$ & $3.82$ & $5.2$\\
\normalfont{\textsuperscript{\ddag}Relative} & $51.54\%$ & $73.46\%$ & $100.00\%$\\[4pt]
\multicolumn{1}{@{}l@{}}{\normalfont{Model = BERT}} & & & \\
\normalfont{Accuracy (\%)} & $1.91$ &  $89.57$ & $90.88$ \\ 
\normalfont{\textsuperscript{\ddag}Relative} & $2.10\%$ & $98.55\%$ & $100.00\%$ \\[4pt]
\normalfont{Size (MB)} & $150.99$ & $301.99$ & $603.98$  \\
\normalfont{\textsuperscript{\ddag}Relative} & $25\%$ & $50\%$ & $100.00\%$\\[4pt]
\normalfont{Latency (ms)} & $2.33$ & $2.79$  & $4.28$ \\
\normalfont{\textsuperscript{\ddag}Relative} & $54.44\%$ & $65.19\%$ & $100.00\%$\\
\bottomrule
\end{tabular}
\end{small}
\end{center}
\vskip -0.2in
\end{table}

We provide absolute and relative model size, accuracy, and deployment latency results for different degrees of quantization for ResNet50 and BERT models in Table~\ref{tab:base}. We designate the 16-bit floating point model which meets MLPerf accuracy threshold as our baseline, for relative results. Quantized models all use the same bit-precision for all their weights and activations. Model size reduction scale linearly with the number of bits while latency reduction captures the complex interactions that real-world execution will encounter. Further software support for heavily quantized operations could further improve the compute latencies.

\subsection{Results}

\begin{table*}[t]
\ra{1.3}
\caption{Results for ResNet50 on ImageNet and BERT on SQuAD. All numbers are percentages relative to the size and latency of a floating point model with 16 bits which can be seen in Table~\ref{tab:base}. For random sensitivity (e.g. uniformed sensitivity) ordering we repeated the experiments five times with different random seeds to ensure a representative result.}
\label{tab:both}
\begin{center}
\begin{small}
\begin{tabular}{@{}rrrlrrlrrlrr@{}}
\toprule
        Model & \multicolumn{5}{c}{ResNet50} & & \multicolumn{5}{c}{BERT} \\
        \cmidrule{2-6} \cmidrule{8-12}
        Accuracy Target &  \multicolumn{2}{c}{99\%} & & \multicolumn{2}{c}{99.9\%} & & \multicolumn{2}{c}{99\%} & & \multicolumn{2}{c}{99.9\%}\\ 
        \cmidrule{2-3} \cmidrule{5-6} \cmidrule{8-9} \cmidrule{11-12}
        ~ & \multicolumn{1}{c}{Size} & \multicolumn{1}{c}{Latency} & & \multicolumn{1}{c}{Size} & \multicolumn{1}{c}{Latency} & & \multicolumn{1}{c}{Size} & \multicolumn{1}{c}{Latency} & & \multicolumn{1}{c}{Size} & \multicolumn{1}{c}{Latency}\\ 
        \midrule
        \multicolumn{1}{@{}l@{}}{Search = Bisection} & & & & &  \\ 
        Random & 50.81\% & 74.11\% & & 51.84\% & 74.10\% & & 59.34\% & 72.27\% & & 92.14\% & 95.04\% \\ 
        $\pm \sigma$ & 0.81\% & 0.56\% & & 1.44\% & 0.51\% & & 2.89\% & 2.06\% & & 3.18\% & 1.65\% \\[3pt]
        $\mathcal{E}_{\text{Hessian}}$ & 50.01\% & 73.98\% & & 50.01\% & 73.98\% & & 72.57\% & 77.61\% & & 81.08\% & 84.65\% \\
        $\mathcal{E}_{\text{N}}$ & 52.00\% & 73.69\% & & 58.94\% & 79.04\% & & 54.77\% & 68.96\% & & 81.42\% & 87.58\% \\ 
        $\mathcal{E}_{\text{QE}}$ & 50.02\% & 73.71\% & & 50.02\% &73.71\% & & 88.20\% & 89.07\% & & 93.75\% & 93.87\% \\[4pt]
        \multicolumn{1}{@{}l@{}}{Search = Greedy } & & & & &  \\
        Random & 49.74\% & 73.05\% & & 49.92\% & 73.30\% & & 53.21\% & 67.45\% & & 69.10\% & 78.51\%\\ 
        $\pm\sigma$ & 0.14\% & 0.28\% & & 0.06\% & 0.18\% & & 4.27\% & 2.44\% & & 5.22\% & 3.77\% \\[3pt] 
        $\mathcal{E}_{\text{Hessian}}$  & \textbf{49.22\%} &  \textbf{72.41\%} & & \textbf{49.86\%} &  \textbf{73.14\%} & & \textbf{49.91\%} & \textbf{65.69\%} & & \textbf{68.40\%} & \textbf{76.60}\% \\
        $\mathcal{E}_{\text{N}}$ & 49.73\% & 72.68\% & & 49.94\% & 73.32\% & & 55.21\% & 69.12\% & & 72.05\% & 81.21\% \\ 
        $\mathcal{E}_{\text{QE}}$  & 49.94\% & 73.37\% & & 49.86\% & 73.14\% & & 70.92\% & 78.08\% & & 78.30\% & 83.97\% \\
\bottomrule
\end{tabular}
\end{small}
\end{center}
\vskip -0.2in
\end{table*}

\begin{figure*}[ht]
\begin{center}
\centerline{\includegraphics[width=1.0\textwidth]{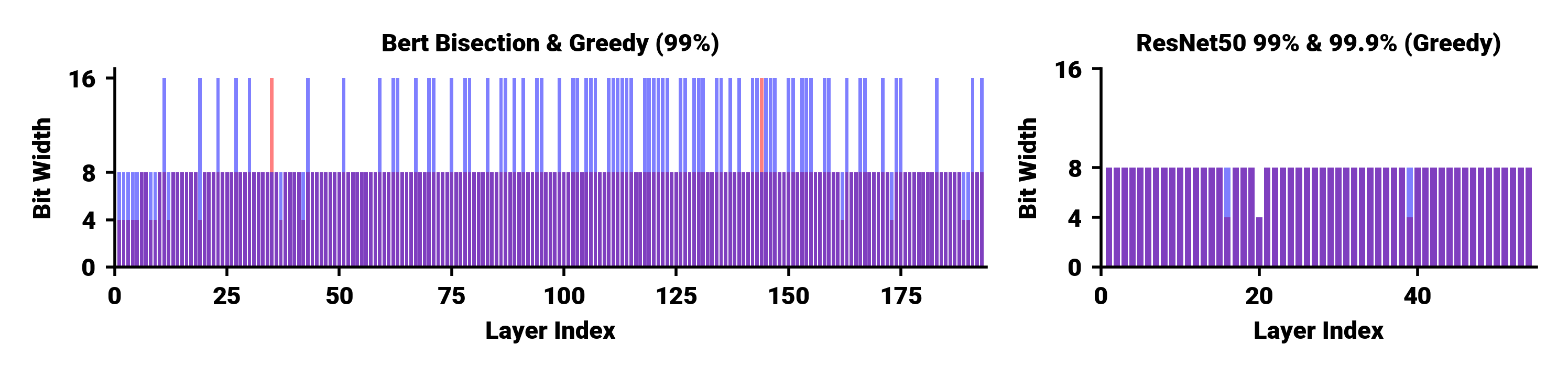}}
\vskip -0.2in
\caption{Illustrating per-layer bit width configurations for BERT and ResNet50. Left highlights when different configurations arise  between bisection (blue) and greedy (red). Due to its sensitivity to ordering, bisection search configures substantially more layers to operate at 16 bits. Right shows the different bit allocation between a $99\%$ (red) and $99.9\%$ (blue) accuracy target for ResNet50 with the differences in bit width allocations arising from a few additional layers quantized to 4 bits.}
\label{fig:comp}
\end{center}
\vskip -0.2in
\end{figure*}

We tabulate our results in Table~\ref{tab:both}, showing the relative model footprint and inference latencies resulting from applying our methods compared to the reference baseline (Table~\ref{tab:base}). We examine the improvement given two target accuracies ($99\%$ and $99.9\%$, with results for 90\% in the appendix Table~\ref{tab:bert}). Hessian-guided greedy search consistently  outperforms all other methods, compressing both models by more than $50\%$ while maintaining a target accuracy of $99\%$ relative to the baseline. The results indicate that even though the models can be compressed by a similar factor, the different compute kernels do not see similar latency benefits. BERT benefits from a larger reduction in latency ($65.69\%$ of the baseline latency) while ResNet50 only reduces to $72.41\%$ of the baseline. More stringent accuracy constraints do not substantially change ResNet50 inference latency, while BERT sees a reduction to $\approx76\%$ relative to the baseline. The changes of latency and model size between the $99\%$ and $99.9\%$ target for the ResNet50 are small meanwhile for the BERT model changes $>10\%$ highlighting the difference between the ease of quantizing these models. Figure~\ref{fig:comp} shows that for ResNet50 models targeting $99\%$ (red) and the $99.9\%$ (blue) accuracy, only a few additional layers can be quantized to 4-bits for a minimal improvement in latency.

\paragraph{Sensitivity Metrics Evaluation}
\begin{figure}[ht]
\begin{center}
\centerline{\includegraphics[width=\columnwidth]{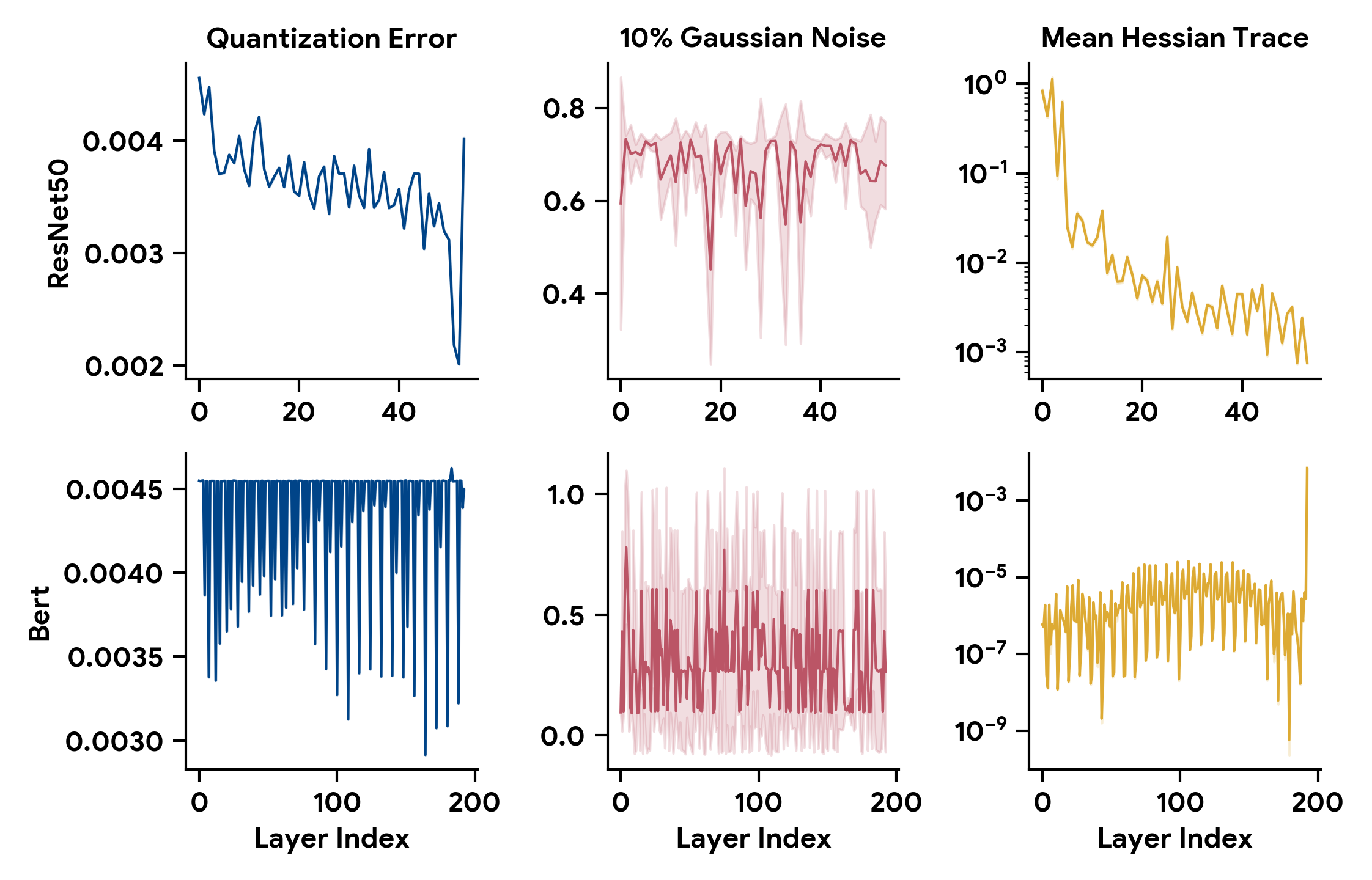}}
\vskip -0.2in
\caption{Sensitivty Metrics for ResNet50 and BERT: quantization Error $\mathcal{E}_{\text{QE}}$, accuracy degradation from noise $\mathcal{E}_{\text{N}}$, and second derivative information $\mathcal{E}_{\text{Hessian}}$. Metrics were obtained over five trials and the shaded areas indicate one standard deviation. The noise based measure has a higher variability compared to both other metrics.}
\label{fig:sens}
\end{center}
\vskip -0.2in
\end{figure}

Figure~\ref{fig:sens} visualizes the layer orderings for ResNet50 and BERT, derived from the three sensitivity metrics. Solid lines denote the average of five trials and shaded areas show the magnitude of one standard deviation. The across trials, there is minimal variance between runs for both QE and the Hessian Trace (not visible in the figure). For ResNet50 both QE and the Hessian Trace indicate that earlier layers are more sensitive, while for Bert there is a repeating structure in the transformer blocks with varying sensitivity. We computed the Levenshtein distance (edit distance) between the resulting orderings of the layers and note that generally all metrics produce orderings which are far from each other, e.g. for ResNet50 the closest pair is $\mathcal{E}_{QE}$ and $\mathcal{E}_{\text{Hessian}}$ with an distance of $48$ (the maximum distance here would be 54).

We first examine how much guidance from the sensitivity metric might be able to improve configuration search. We create a random ordering for the layers and implement both the bisection and greedy search on these. The results are evaluated over 5 trials with the expected relative latency and model compression and the standard deviation over these trials, reported in Table~\ref{tab:both}. For ResNet50 all sensitivity measures exhibit similar latency and model size reductions as the uninformed baseline (random), around 50\% model compression for both accuracy targets. On the other hand BERT shows improvements from an informative search, with the random baseline achieving a model footprint that is $\sim53\%$ of the baseline for a target accuracy of $99\%$. Both QE ($\mathcal{E}_{\text{QE}}$) and the Noise-based sensitivity estimates underperform the random `guidance', with QE compressing model footprint to $\sim78\%$ of the baseline and Noise-based sensitivity achieving footprints that were $\sim72\%$ of the baseline. However, the inconsistent results from the uninformed guidance (see large standard deviation in Table~\ref{tab:both} and Figure~\ref{fig:sens} in the appendix), indicate that this metric might be unsuitable deployment scenarios. Using the Hessian Trace for guidance, consistently outperforms the random baseline by 1-3\% for all accuracy targets when using the greedy search. Indicating tangible benefits with additional computation.

\paragraph{Search Algorithm Evaluation}
For both models (ResNet50 and BERT) Greedy search results in a more performant quantized model across sensitivity metrics (including the uninformed baseline). Although quantizing ResNet50 does not offer dramatic benefits. The benefits of the greedy approach become more apparent on examining the quantization results for BERT (Table~\ref{tab:both}), where a Greedy search improves model compression compared to bisection by $10\%$ on average. Of the examined approaches, only models quantized using the greedy search consistently outperform the baselines. Figure~\ref{fig:comp} outlines how different layers are configured across some of the studied search methods. Greedy search for BERT (red in Figure~\ref{fig:comp} left) more aggressively configures layers to operate at 4b whereas bisection search leaves more layers at 16b for a $99\%$ target accuracy.

\paragraph{Comparison to Prior Work}
Figure~\ref{fig:over} contextualizes our results with respect to other work. Most other PTQ efforts do not meet accuracies acceptable for easy deployment and typically use a single bit-width configurations across the entire model. For quantizing ResNet50, \citet{wu2020integer} approaches the $99\%$ accuracy threshold, however since they only apply single bit-configuration across the model they are unable to increase the precison of critical layers to to achieve the target accuracy. For quantizing BERT, only the work outlined in~\cite{shen2020q} provides competitive accuracy. Again, our multi-precision method delivers finer control over bit-width configuration enabling higher model accuracy at comparable model size.

\section{Conclusions}

In this paper we developed a practical quantization pipeline for efficient deployment of floating point NN using quantized tensor vector multiplications of various bit-widths. We use PTQ, which calibrates the quantizer scales on data and subsequently uses back-propagation to refine them, without changing any of the original model parameters. We are able to determine a suitable quantization configuration that can meet previously set accuracy expectations while still delivering model compression. This is crucial for practical deployment of large-scale models that can leverage the full quantization potential of modern ML accelerator hardware. We evaluate different sensitivity metrics to guide configuration search, and showed the benefits resulting from using a Hessian-based metric. We evaluate two guided search algorithms: bisection and greedy search that can rapidly search the large configuration space guided by the sensitivity metrics. The resulting configurations consistently outperformed state-of-the art PTQ, consistently meeting a $99\%$ accuracy requirement with models compressed to $\le50\%$ of the floating point baseline for both ResNet50 (evaluated on ImageNet) and BERT (evaluated on SQUAD). We demonstrate that most of the benefits for mixed-precision quantization can be obtained without any sensitivity guidance while using a progressive Greedy approach.

\paragraph{Acknowledgement} The authors gratefully acknowledge the valuable and profound feedback from Daniel Finchelstein and Naveen Kumar.

\bibliography{references}
\bibliographystyle{icml2022}

%%%%%%%%%%%%%%%%%%%%%%%%%%%%%%%%%%%%%%%%%%%%%%%%%%%%%%%%%%%%%%%%%%%%%%%%%%%%%%%
%%%%%%%%%%%%%%%%%%%%%%%%%%%%%%%%%%%%%%%%%%%%%%%%%%%%%%%%%%%%%%%%%%%%%%%%%%%%%%%
% APPENDIX
%%%%%%%%%%%%%%%%%%%%%%%%%%%%%%%%%%%%%%%%%%%%%%%%%%%%%%%%%%%%%%%%%%%%%%%%%%%%%%%
%%%%%%%%%%%%%%%%%%%%%%%%%%%%%%%%%%%%%%%%%%%%%%%%%%%%%%%%%%%%%%%%%%%%%%%%%%%%%%%
\newpage
\appendix
\onecolumn

\section{Results for 90\% Accuracy Target}

\begin{table*}[h!]
\ra{1.3}
\caption{Results for for ResNet50 on ImageNet and BERT on SQuAD given a 90\% accuracy target.  All numbers are percentages relative to the size and latency of a floating point model with 16 bits which can be seen in Table~\ref{tab:base}.}
\label{tab:bert}
\begin{center}
\begin{small}
\begin{tabular}{@{}rrrlrr@{}}
\toprule
        Model &  \multicolumn{2}{c}{ResNet50} & & \multicolumn{2}{c}{BERT} \\ 
        \cmidrule{2-3} \cmidrule{5-6}
        ~ & \multicolumn{1}{c}{Size} & \multicolumn{1}{c}{Latency} & & \multicolumn{1}{c}{Size} & \multicolumn{1}{c}{Latency}\\ 
        \midrule
        \multicolumn{1}{@{}l@{}}{Search = Bisection} & & & & &  \\         
        Random & 49.80\% & 73.48\% & & 49.17\% & 65.09\%\\
        $\pm\sigma$ & 0.71\% & 0.91\% & & 0.39\% & 0.17\% \\[3pt]
        $\mathcal{E}_{\text{Hessian}}$ & 45.69\% & 73.32\% & & 48.87\% & 65.49\%\\ 
        $\mathcal{E}_{\text{N}}$ & 51.87\% & 71.98\% & & 48.52\% & 64.82\% \\ 
        $\mathcal{E}_{\text{QE}}$ & 46.68\% & 72.80\% & & 48.96\% & 65.28\% \\[4pt]         
        \multicolumn{1}{@{}l@{}}{Search = Greedy} & & & & &  \\
        Random & 47.61\% & 71.19\% & &  46.79\% & 63.94\% \\ 
        $\pm\sigma$ & 1.19\% & 0.87\% & &  0.75\% & 0.32\% \\[3pt]
        $\mathcal{E}_{\text{Hessian}}$ &  \textbf{44.17\%} &  \textbf{70.83\%} & & \textbf{45.92\%} & \textbf{63.71\%} \\        
        $\mathcal{E}_{\text{N}}$ & 49.46\% & 70.86\% & & 46.53\% & 63.86\% \\
        $\mathcal{E}_{\text{QE}}$ & 45.30\% & 71.28\% & & 46.27\% & 63.84\%\\
\bottomrule
\end{tabular}
\end{small}
\end{center}
\vskip -0.2in
\end{table*}

\end{document}